\documentclass{article}
\usepackage{ijcai18}

\usepackage{times}
\usepackage{xcolor}
\usepackage{soul}
\usepackage[utf8]{inputenc}
\usepackage[small]{caption}
\usepackage{amsmath}

\usepackage[ruled]{algorithm2e}
\usepackage{multirow} 
\usepackage{graphicx}
\usepackage{subfig}
\usepackage{threeparttable}

\usepackage{caption}
\usepackage{algorithmic}

\newtheorem{theorem}{Theorem}
\DeclareMathOperator*{\argmax}{\arg\!\max}
\title{Self-weighted Multiple Kernel Learning for Graph-based \\Clustering and Semi-supervised Classification}

\author{
Zhao Kang$^{1*}$, 
Xiao Lu$^1$, 
Jinfeng Yi$^2$, 
Zenglin Xu$^1$\thanks{Corresponding author.}, 
\\ 
$^1$ SMILE Lab, School of Computer Science and Engineering\\
University of Electronic Science and Technology of China, Sichuan 611731, China \\
$^2$  JD AI Research, Beijing 100101, China \\
zkang@uestc.edu.cn,
nbshawnlu@hotmail.com,
jinfengyi.ustc@gmail.com,
zenglin@gmail.com
}
\begin{document}

\maketitle

\begin{abstract}
Multiple kernel learning (MKL) method is generally believed to perform better than single kernel method. However, some empirical studies show that this is not always true: the combination of multiple kernels may even yield an even worse performance than using a single kernel. There are two possible reasons for the failure: (i) most existing MKL methods assume that the optimal kernel is a linear combination of base kernels, which may not hold true; and (ii) some kernel weights are inappropriately assigned due to noises and carelessly designed algorithms. In this paper, we propose a novel MKL framework by following two intuitive assumptions: (i) each kernel is a perturbation of the consensus kernel; and (ii) the kernel that is close to the consensus kernel should be assigned a large weight. Impressively, the proposed method can automatically assign an appropriate weight to each kernel without introducing additional parameters, as existing methods do. The proposed framework is integrated into a unified framework for graph-based clustering and semi-supervised classification. We have conducted experiments on multiple benchmark datasets and our empirical results verify the superiority of the proposed framework.

\end{abstract}

\section{Introduction}
As a principled way of introducing non-linearity into linear models, kernel methods have been widely applied in many machine learning tasks  \cite{hofmann2008kernel,xu2010simple}. Although improved performance has been reported in a wide variety of problems, the kernel methods require the user to select and tune a single pre-defined kernel. This is not user-friendly since the most suitable kernel for a specific task is usually challenging to decide. Moreover, it is time-consuming and impractical to exhaustively search from a large pool of candidate kernels. Multiple kernel learning (MKL) was proposed to address this issue as it offers an automatical way of 
learning an optimal combination of distinct base kernels \cite{xu2009extended}. Generally speaking, MKL method should yield a better performance than that of single kernel approach. 

A key step in MKL is to assign a reasonable weight to each kernel according to its importance. One popular approach considers a weighted combination of candidate kernel matrices, leading to a convex quadratically constraint quadratic program. However, this method over-reduces the feasible set of the optimal kernel, which may lead to a less representative solution. In fact, these MKL algorithms sometimes fail to outperform single kernel methods or traditional non-weighted kernel approaches~\cite{yu20102,gehler2009feature}. 
Another issue is the inappropriate weights assignment. 
Some attempts aim to learn the local importance of features by assuming that samples may vary locally~\cite{gonen2008localized}. However, they induce more complex computational problems. 

To address these issues, in this paper, we model the differences among kernels by following two intuitive assumptions: (i) each kernel is a perturbation of the consensus kernel; and (ii) the kernel that is close to the consensus kernel should receive a large weight. As a result, instead of enforcing the optimal kernel being a linear combination of predefined kernels, this approach allows the most suitable kernel to reside in some kernels' neighborhood. And our proposed method can assign an optimal weight for each kernel automatically without introducing an additive parameter as existing methods do. 

Then we combine this novel weighting scheme with graph-based clustering and semi-supervised learning (SSL). Due to its effectiveness in similarity graph construction, graph-based clustering and SSL have shown impressive performance \cite{nie2017multi,kang2017twin}. Finally, a novel multiple kernel learning framework for clustering and semi-supervised learning is developed.

In summary, our main contributions are two-fold:
\begin{itemize}
\itemsep-.2em 
\item We proposed a novel way to construct the optimal kernel and assign weights to base kernels. Notably, our proposed method can find a better kernel in the neighborhood of candidate kernels. This weight is a function of kernel matrix, so we do not need to introduce an additional parameter as existing methods do. This also eases the burden of solving the constraint quadratic program.
\item A unified framework for clustering and SSL is developed. It seamlessly integrates the components of graph construction, label learning, and kernel learning by incorporating the graph structure constraint. This allows them to negotiate with each other to achieve overall optimality. Our experiments on multiple real-world datasets verify the effectiveness of the proposed framework. 
\end{itemize}

\section{Related Work}
In this section, we divide the related work into two categories, namely graph-based clusteirng and SSL, and paremeter-weighted multiple kernel learning.

\subsection{Graph-based Clustering and SSL}
Graph-based clustering \cite{ng2002spectral,yang2017discrete} and SSL \cite{zhu2003semi} have been popular for its simple and impressive performance. The graph matrix to measure the similarity of data points is crucial to their performance and there is no satisfying solution for this problem. Recently, automatic learning graph from data has achieved promising results. One approach is based on adaptive neighbor idea, i.e., $s_{ij}$ is learned as a measure of the probability that $X_i$ is neighbor of $X_j$. Then $S$ is treated as graph input to do clustering \cite{nie2014clustering,huang2018adaptive} and SSL \cite{nie2017multi}. Another one is using the so-called self-expressiveness property, i.e., each data point is expressed as a weighted combination of other points and this learned weight matrix behaves like the graph matrix. Representative work in this category include \cite{huang2015new,li2015learning,kang2018unified}. These methods are all developed in the original feature space. To make it more general, we develop our model in kernel space in this paper. Our purpose is to learn a graph with exactly $c$ number of connected components if the data contains $c$ clusters or classes. In this work, we will consider this condition explicitly.
\subsection{Parameter-weighted Multiple Kernel Learning}
\label{submkl}
  It is well-known that the performance of kernel method crucially depends on the kernel function as it intrinsically specifies the feature space. MKL is an efficient way for automatic kernel selection and embedding different notions of similarity \cite{kang2017kernel}. It is generally formulated as follows:
\begin{equation}
\min_\theta \quad f(K) \quad  s.t.\quad K=\sum_i \theta_i H^i,\hspace{.1cm} \sum_i (\theta_i)^q=1,\hspace{.1cm}\theta_i\ge 0,
\label{mkl}
\end{equation}
where $f$ is the objective function, $K$ is the consensus kernel, $H^i$ is our artificial constructed kernel, $\theta_i$ represents the weight for kernel $H^i$, $q>0$ is used to smoothen the weight distribution. Therefore, we frequently sovle $\theta$ and tune $q$. Though this approach is widely used, it still suffers the following problems. First, the linear combination of base kernels over reduces the feasible set of optimal kernels, which could result in the learned kernel with limited representation ability. Second, the optimization of kernel weights may lead to inappropriate assignments due to noise and carelessly designed algorithms. Indeed, contrary to the original intention of MKL, this approach sometimes obtains lower accuracy than that of using equally weighted kernels or merely single kernel method. This will hinder the practical use of MKL. This phenomenon has been observed for many years \cite{cortes2009can} but rarely studied. Thus, it is vital to develop some new approaches.

\section{Methodology}
\subsection{Notations}
 Throughout the paper, all the matrices are written as uppercase. For a matrix $X$, its $ij$-th element and $j$-th column is denoted as $x_{ij}$ and $X_j$, respectively. The trace of $X$ is denoted by $Tr(X)$. The $i$-th kernel of $X$ is written as $H^i$. The $p$-norm of vector $x$ is represented by $\|x\|_p$. The Frobenius norm of matrix $X$ is denoted by $\|X\|_F$. $I$ is an identity matrix with proper size. $S\geq 0$ means all entries of $S$ are nonnegative.
 
\subsection{Self-weighted Multiple Kernel Learning}
Aforementioned self-expressiveness based graph learning method can be formulated as:
\begin{equation}
\min_{S} \|X-XS\|_F^2+\gamma \|S\|_F^2 \hspace{.2cm} s.t. \hspace{.2cm} S\geq 0,
\label{global}
\end{equation} 
where $\gamma$ is the trade-off parameter. To recap the powerfulness of kernel method, we extend Eq. (\ref{global}) to its kernel version by using kernel mapping $\phi$. According to the kernel trick $K(x,z)=\phi(x)^T\phi(z)$, we have
\begin{equation}
\begin{split}
&\min_{S}\hspace{.1cm} \|\phi(X)-\phi(X)S\|_F^2+\gamma \|S\|_F^2,\\
\Longleftrightarrow
&\min_{S} \hspace{.1cm} Tr(K-2KS+S^T KS)+\gamma \|S\|_F^2\\
&\quad s.t. \quad S\ge0
\label{kernel}
\end{split}
\end{equation}
Ideally, we hope to achieve a graph with exactly $c$ connected components if the data contain $c$ clusters or classes. In other words, the graph $S$ is block diagonal with proper permutations. It is straightforward to check that $S$ in Eq. (\ref{kernel}) can hardly satisfy to such a constraint condition. 

If the similarity graph matrix $S$ is nonnegative, then the Laplacian matrix $L=D-S$, where $D$ is the diagonal degree matrix defined as $d_{ii}=\sum_j s_{ij}$, associated with $S$ has an important property as follows \cite{mohar1991laplacian}
\begin{theorem}
The multiplicity $c$ of the eigenvalue 0 of the Laplacian matrix $L$ is equal to the number
of connected components in the graph associated with $S$.
\end{theorem}
Theorem 1 indicates that if $rank(L)=n-c$, then the constraint on $S$ will be held. Therefore, the problem (\ref{kernel}) can be rewritten as:
\begin{equation}
\begin{split}
&\min_{S} \hspace{.1cm} Tr(K-2KS+S^T KS)+\gamma \|S\|_F^2\\
&\quad s.t. \quad S\ge0,\hspace{.1cm} rank(L)=n-c
\end{split}
\label{rank}
\end{equation}
Suppose $\sigma_i(L)$ is the $i$-th smallest eigenvalue of $L$. Note that $\sigma_i(L)\ge 0$ because $L$ is positive semi-definite.
The problem (\ref{rank}) is equivalent to the following problem for a large enough $\alpha$:
\begin{equation}
\begin{split}
&\min_{S} \hspace{.1cm} Tr(K-2KS+S^T KS)+\gamma \|S\|_F^2
+\alpha \sum\limits_{i=1}^c \sigma_i(L)\\
&\quad s.t. \quad S\ge0
\end{split}
\label{rank2}
\end{equation}
According to the Ky Fan's Theorem \cite{fan1949theorem}, we have:
\begin{equation}
\sum\limits_{i=1}^c \sigma_i(L)=\min_{P\in\mathcal{R}^{n\times c},P^TP=I} Tr(P^TLP)
\end{equation}
$P$ can be cluster indicator matrix or label matrix. Therefore, the problem (\ref{rank2}) is further equivalent to the following problem
\begin{equation}
\begin{split}
&\min_{S,P}Tr(K-2KS+S^TKS)+\gamma \|S\|_F^2+\alpha Tr(P^TLP)\\
&\quad s.t.\quad P^TP=I,\quad S\geq 0
\end{split}
\label{equi}
\end{equation}
This problem (\ref{equi}) is much easier to solve compared with the rank constrained problem (\ref{rank}). We name this model as \textbf{K}ernel-based \textbf{G}raph \textbf{L}earning (KGL). Note that this model's input is kernel matrix $K$. It is generally recognized that its performance is largely determined by the choice of kernel. Unfortunately, the most suitable kernel for a particular task is often unknown in advance. Although MKL as in Eq. (\ref{mkl}) can be applied to resolve this issue, it is still not satisfying as we discussed in subsection \ref{submkl}. 

In this work, we design a novel MKL strategy. It is based on the following two intuitive assumptions: 1) each kernel is a perturbation of the consensus kernel, and 2) the kernel that is close to the consensus kernel should receive a large weight. Motivated by these, we can have the following MKL form:
\begin{equation}
\min_K \sum_i w_i\|H^i-K\|_F^2 
\label{self}
\end{equation} 
and
\begin{equation}
w_i=\frac{1}{2\|H^i-K\|_F}
\label{solvew}
\end{equation}
We can see that $w_i$ is dependent on the target variable $K$, so it is not directly available. But $w_i$ can be set to be stationary, i.e., after obtaining $K$, we update $w_i$ correspondingly \cite{nie2017self}. Instead of enforcing the optimal kernel being a linear combination of candidate kernels as in Eq. (\ref{mkl}), Eq. (\ref{self}) allows the most suitable kernel to reside in some kernels' neighborhood \cite{liu2009learning}. This enhances the representation ability of the learned optimal kernel \cite{liu2017optimal,liu2013adaptive}. Furthermore, we don't introduce an additive hyperparameter $\theta$, which often leads to a quadratic program. The optimal weight $w_i$ for each kernel $H^i$ is directly calculated according to kernel matrices. Then our Self-weighted Multiple Kernel Learning (SMKL) framework can be formulated as:
\begin{equation}
\begin{split}
&\min_{S,P,K}Tr(K-2KS+S^TKS)+\gamma \|S\|_F^2+\alpha Tr(P^TLP)\\
&+\beta \sum_{i=1}^r w_i\|H^i-K\|_F^2\quad s.t.\quad P^TP=I,\hspace{.1cm} S\geq 0.
\end{split}
\label{obj}
\end{equation}
This model enjoys the following properties:
\begin{enumerate}
\itemsep-.2em 
\item This unified framework sufficiently considers the negotiation between the process of learning the optimal kernel and that of graph/label learning. By iteratively updating $S$, $P$, $K$, they can be repeatly improved.
\item By treating the optimal kernel as a perturbation of base kernels, it effectively enlarges the region from which an optimal kernel can be chosen, and therefore is in a better position than the traditional ones to identify a more suitable kernel. 
\item The kernel weight is directly calculated from kernel matrices. Therefore, we avoid solving quadratic program.
\end{enumerate}
To see the effect of our proposed MKL method, we need to examine the approach with traditional kernel learning. For convenience, we denote it as Parameterized MKL (PMKL). It can be written as:  
\begin{equation}
\begin{split}
&\min_{S,P,\theta}Tr(K-2KS+S^TKS)+\gamma \|S\|_F^2+\alpha Tr(P^TLP)\\
&\quad s.t.\quad  K=\sum\limits_{i=1}^r \theta_i H^i,\hspace{.1cm}\sum\limits_{i=1}^r \sqrt{\theta_i}=1,\hspace{.1cm} \theta_i\ge 0,\\ 
&\quad\quad\quad   P^TP=I,\hspace{.1cm} S\geq 0.
\end{split}
\label{pmkl}
\end{equation}

\subsection{Optimization}
We divide the problem in Eq. (\ref{obj}) into three subproblems, and develop an alternative and iterative algorithm to solve them.

For $S$, we fix $P$ and $K$. The problem in Eq. (\ref{obj}) becomes:
\begin{equation}
\begin{split}
\min_{S}&\hspace{.1cm}  Tr(-2KS+S^TKS)+\gamma \|S\|_F^2+\alpha Tr(P^TLP),\\
&s.t.\quad S\geq 0.
\end{split}
\end{equation}
Based on $\sum_{ij}\frac{1}{2}\|P_{i,:}-P_{j,:}\|_2^2s_{ij}=Tr(P^TLP)$, we can equivalently solve the following problem for each sample:
\begin{equation}
-2K_{i,:}S_i+S_i^T K S_i+\gamma S_i^TS_i+\frac{\alpha}{2} G_i^T S_i,
\label{solvez}
\end{equation}
where $G_i\in\mathcal{R}^{n\times 1}$ with $g_{ij}=\|P_{i,:}-P_{j,:}\|_2^2$. By setting its first derivative w.r.t. $S_i$ to be zero, we obtain:
\begin{equation}
S_i=(\gamma I+K)^{-1}(K_{i,:}-\frac{\alpha G_i}{4}).
\label{solvez}
\end{equation}
Thus $S$ can be achieved in parallel. 

For $K$, we fix $S$ and $P$. The problem in Eq. (\ref{obj}) becomes:
\begin{equation}
\min_{K}Tr(K-2KS+S^TKS)+\beta \sum_i w_i\|H^i-K\|_F^2
\end{equation}

Similar to (\ref{solvez}), it yields:
\begin{equation}
K=\frac{2S^T-SS^T-I+2\beta\sum_i w_iH^i}{2\beta \sum_i  w_i}.
\label{solvek}
\end{equation}
From Eq. (\ref{solvek}) and Eq. (\ref{solvez}), we can observe that $S$ and $K$ are seamlessly coupled, hence they are allowed to negotiate with each other to achieve better results.

For $P$, we fix $S$ and $K$.  The problem in Eq. (\ref{obj}) becomes:
\begin{equation}
\min_{P}Tr(P^TLP)\quad s.t.\quad P^TP=I.
\end{equation} 
The optimal solution $P$ is the $c$ eigenvectors of $L$ corresponding to the $c$ smallest eigenvalues.

\subsection{Extend to Semi-supervised Classification}
Model (\ref{obj}) also lends itself to semi-supervised classification. Graph construction and label inference are two fundamental stages in SSL. Solving two separate problems only once is suboptimal since label information is not exploited when it learns the graph. SMKL unifies these two fundamental components into a unified framework. Then the given labels and estimated labels will be utilized to build the graph and to predict the unknown labels. 

Based on a similar approach, we can reformulate SMKL for semi-supervised classification as:
\begin{equation}
\begin{split}
\min_{S,P,K}&\hspace{.1cm}Tr(K-2KS+S^TKS)+\gamma \|S\|_F^2+\alpha Tr(P^TLP)\\
&+\beta \sum_i w_i\|H^i-K\|_F^2\quad  s.t.\quad S\geq 0,\hspace{.1cm} P_l=Y_l
\end{split}
\label{ssl}
\end{equation}
where $Y_l=[y_1,\cdots,y_l]^T$ denote the label matrix and $l$ is the number of labeled points. $y_i\in\mathcal{R}^{c\times 1}$ is one-hot and $y_{ij}=1$ indicates that the $i$-th sample belongs to the $j$-th class. $P=[P_l; P_u]=[Y_l; P_u]$, where the unlabeled $u$ points in the back. (\ref{ssl}) can be solved in the same procedure as (\ref{obj}), the difference lies in updating $P$. 

To solve $P$, we take the derivative of (\ref{ssl}) with respect to $P$, we have $LP=0$, i.e.,  
\[
\begin{bmatrix}
L_{ll} &L_{lu}\\
L_{ul} & L_{uu}
\end{bmatrix}  
\begin{bmatrix}
Y_{l} \\
P_{u}
\end{bmatrix}  
=0.
\]
It yields:
\begin{equation}
P_u=-L_{uu}^{-1}L_{ul}Y_l.
\label{pssl}
\end{equation}

Finally, the class label for unlabeled points could be assigned according to the following decision rule:

\begin{equation}
 y_i=\argmax_j P_{ij}.
\end{equation}

\begin{algorithm}[htbp]
\caption{The Proposed Framework SMKL}
\label{alg2}
\small
 {\bfseries Input:} Kernel matrices $\{H^i\}_{i=1}^r$, parameters $\alpha$, $\beta$, $\gamma$.\\
{\bfseries Initialize:} Random matrix $S$, $K=\sum_i H^i/r$.\\
 {\bfseries REPEAT}
\begin{algorithmic}[1]
\STATE  For each $i$, update the $i$-th column of $S$ according to (\ref{solvez}).
\STATE Calculate $K$ by (\ref{solvek}).
\STATE Update $w$ by (\ref{solvew}).
\STATE For clustering, calculte $P$ as the $c$ smallest eigenvectors of $L=D-S$ correspond to the $c$ smallest eigenvalues. For SSL, calculte $P$ according to (\ref{pssl}).
\end{algorithmic}
\textbf{ UNTIL} {stopping criterion is met.}
\end{algorithm}
\begin{table}[!htbp]
\centering
\normalsize
\caption{Statistics of the data sets}
\label{data}
\renewcommand{\arraystretch}{1.}
\begin{tabular}{|l|c|c|c|}
\hline
&\textrm{\# instances}&\textrm{\# features}&\textrm{\# classes}\\\hline
\textrm{YALE}&165&1024&15\\\hline
\textrm{JAFFE}&213&676&10\\\hline
\textrm{YEAST}&1484&1470&10\\\hline
\textrm{TR11}&414&6429&9\\\hline
\textrm{TR41}&878&7454&10\\\hline
\textrm{TR45}&690&8261&10\\\hline
\end{tabular}
\end{table}

\section{Clustering }
In this section, we conduct clustering experiments to demonstrate the efficacy of our method.
\subsection{Data Sets}
We implement experiments on six publicly available data sets. We summarize the information of these data sets in Table \ref{data}. In specific, the first two data sets YALE and JAFFE consist of face images. YEAST is microarray data set. Tr11, Tr41, and Tr45 are derived from NIST TREC Document Database. 

We design 12 kernels. They are: seven Gaussian kernels of the form $K(x,y)=exp(-\|x-y\|_2^2/(td_{max}^2))$, where $d_{max}$ is the maximal distance between samples and $t$ varies over the set $\{0.01, 0.05, 0.1, 1, 10, 50, 100\}$; a linear kernel $K(x,y)=x^T y$; four polynomial kernels $K(x,y)=(a+x^T y)^b$ with $a\in\{0,1\}$ and $b\in\{2,4\}$. Besides, all kernels are rescaled to $[0,1]$ by dividing each element by the largest pairwise squared distance. 

\begin{figure}[!htpb]
\centering
\subfloat[$\beta=10$\label{yale}]{\includegraphics[width=.42\textwidth]{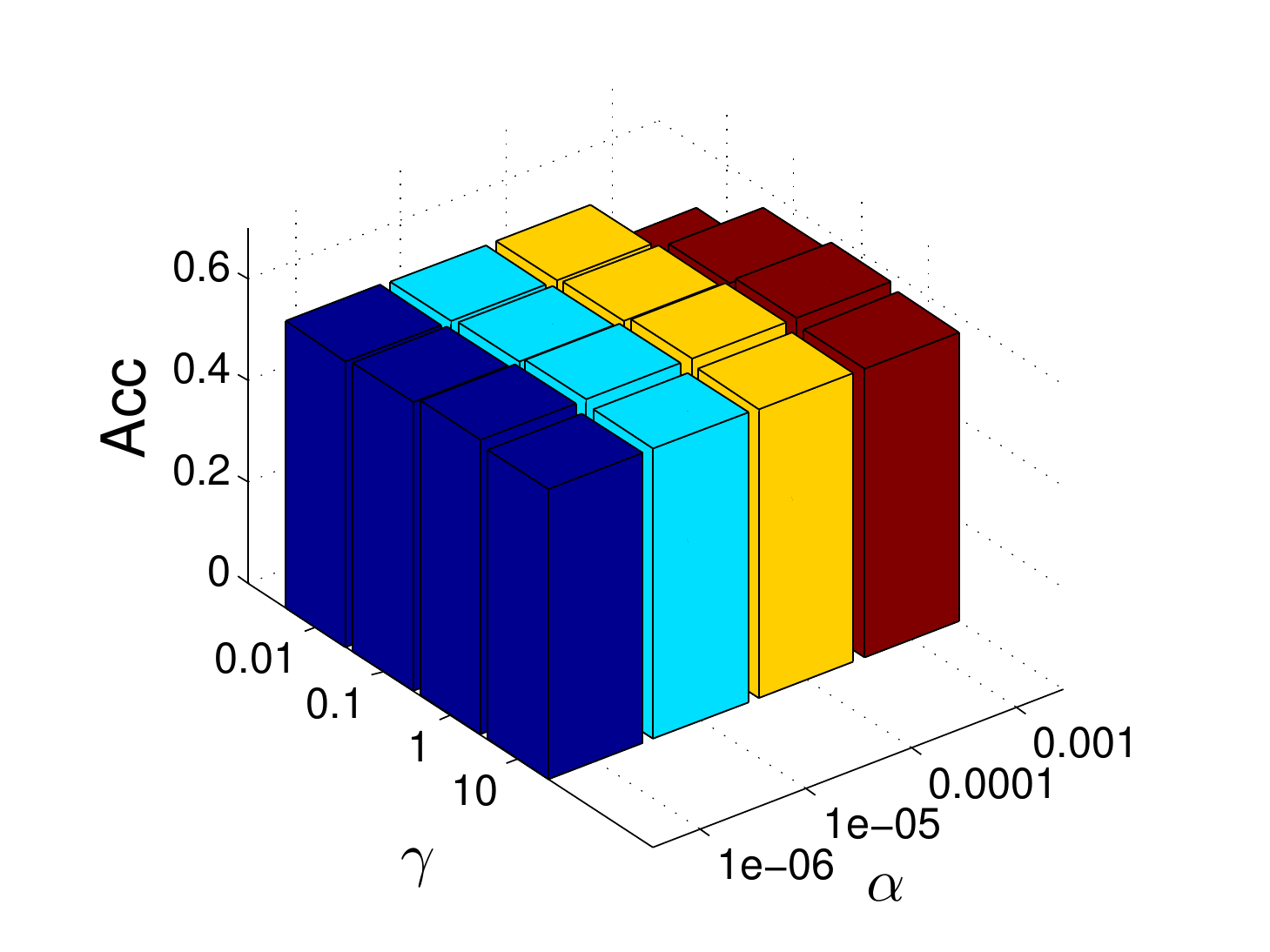}}\\
\subfloat[$\beta=100$\label{yale}]{\includegraphics[width=.42\textwidth]{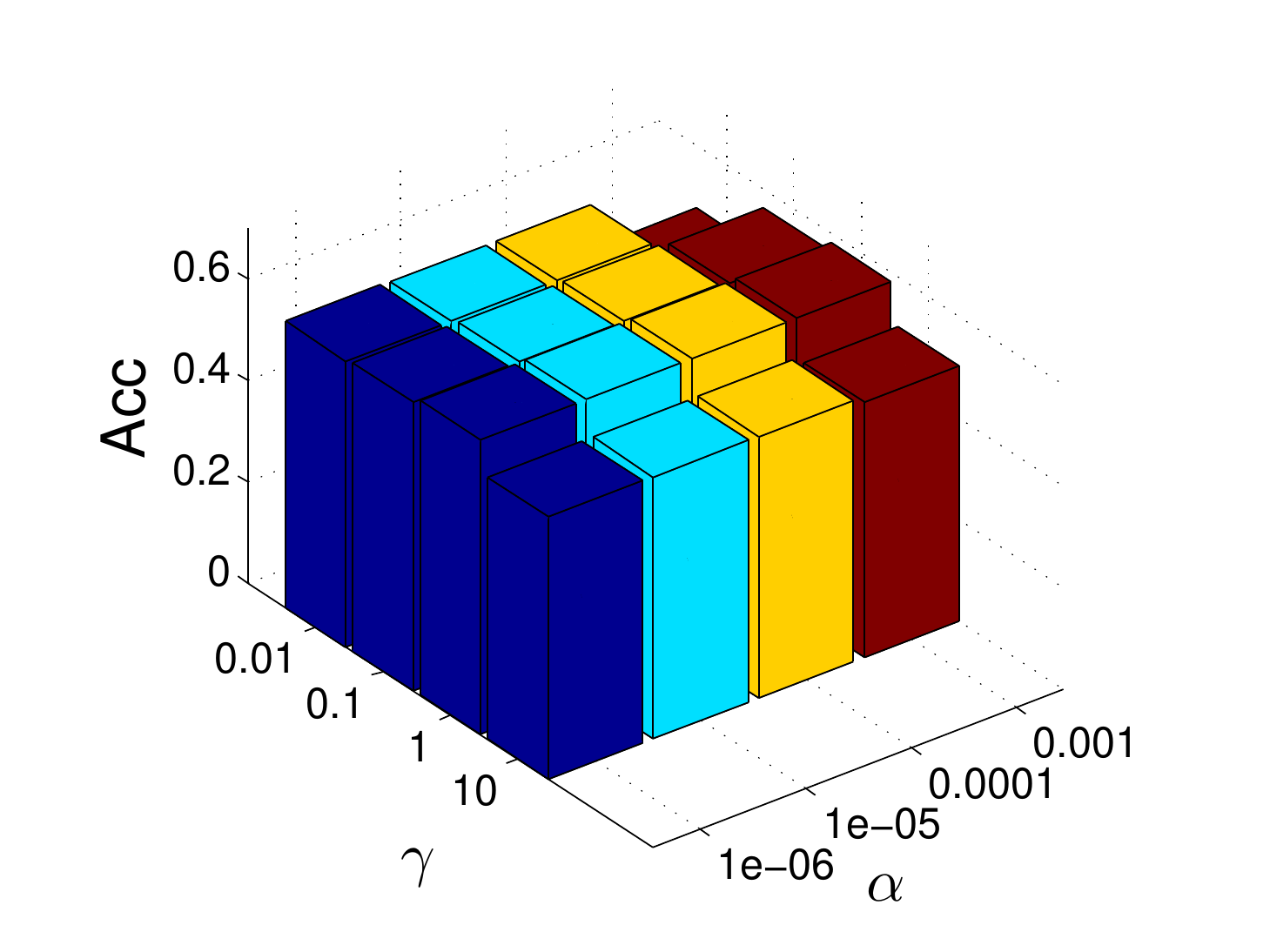}}
\caption{The influence of parameters on accuracy of YALE data set.}
\label{yaleacc}
\end{figure}

\begin{table*}[htbp]
\begin{center}
\tiny
\setlength{\tabcolsep}{1pt}
\renewcommand{\arraystretch}{.9}
\resizebox{.9\textwidth}{!}{
\begin{tabular}{ |l|c|c|c|c|c|c|c|c|c| }
\hline
Data\ \  &\ Metric\ \ &\ SC\ \ &\ \ SSR\ \ & \ \ MKKM\ \  &\ RMKKM\ &\ \ AASC\ \  &\ \ KGL\ \  &\ \ PMKL\ \  &\ \ SMKL\ \  \\

\hline

\multirow{2}{4em}{YALE} & Acc &0.4942&0.5455&0.4570&0.5218&0.4064&0.5549&0.5605&\textbf{0.6000}\\ 
\cline{2-10}
& NMI & 0.5292&0.5726& 0.5006&0.5558& 0.4683&0.5498&0.5643&\textbf{0.6029}\\ 
\hline
\multirow{2}{4em}{JAFFE} & Acc & 0.7488&0.8732& 0.7455& 0.8707&0.3035&0.9877&0.9802& \textbf{0.9906}\\ 
\cline{2-10}
&NMI & 0.8208&0.9293& 0.7979& 0.8937&0.2722&0.9825&0.9806& \textbf{0.9834}\\ 
\hline
\multirow{2}{4em}{YEAST} & Acc &0.3555&0.2999&0.1304&0.3163&0.3538&0.3892&0.3952& \textbf{0.4326}\\ 
\cline{2-10}
&NMI &0.2138&0.1585&0.1029&0.2071&0.2119&0.2315&0.2361& \textbf{0.2652}\\ 
\hline
\multirow{2}{4em}{TR11} & Acc &0.5098&0.4106& 0.5013 &0.5771&0.4715&0.7425&0.7485& \textbf{0.8309}\\ 
\cline{2-10}
& NMI &0.4311&0.2760&0.4456&0.5608&0.3939&0.6000&0.6137&\textbf{0.7167}\\ 
\hline
\multirow{2}{4em}{TR41} & Acc &0.6352&0.6378&0.5610& 0.6265&0.4590&0.6942 &0.6724&\textbf{0.7631}\\ 
\cline{2-10}
& NMI &0.6133&0.5956&0.5775& 0.6347&0.4305&0.6008&\textbf{0.6500}&0.6148\\ 
\hline
\multirow{2}{4em}{TR45}& Acc &0.5739&0.7145& 0.5846&0.6400 & 0.5264&0.7425&0.7468&\textbf{0.7536}\\ 
\cline{2-10}
& NMI &0.4803&0.6782&0.5617& 0.6273&0.4190&0.6824&\textbf{0.7523}&0.6965\\ 
\hline
\end{tabular}}
\end{center}
\caption{Clustering results on benchmark data sets. The best results are in bold font.\label{clusterres}}
\end{table*}

\subsection{Comparison Methods}
We compare with a number of single kernel and multiple kernel learning based clustering methods. They include: Spectral Clustering (SC) \cite{ng2002spectral}, Simplex Sparse Representation (SSR) \cite{huang2015new}, Multiple Kernel k-means (MKKM) \cite{huang2012multiple}, Affinity Aggregation for Spectral Clustering (AASC) \cite{huang2012affinity}, Robust Multiple Kernel k-means (RMKKM)\footnote{https://github.com/csliangdu/RMKKM} \cite{du2015robust}. Among them, SC, SSR, AASC are graph based methods, while the others are k-means variants. Although there are much recent work focused on multiview \cite{huang2018} and deep learning based clustering \cite{peng2016deep}, they are not our focus in this paper.
\subsection{Performance Evaluation}
SMKL is compared with other techniques. We show the clustering results in terms of accuracy (Acc), NMI in Table \ref{clusterres}. For SC and KGL, we report its best performance achieved from those 12 kernels. It can clearly be seen that SMKL achieves the best performance in most cases. Compared to PMKL, SMKL\footnote{https://github.com/sckangz/IJCAI2018} works better. As shown in Eq. (\ref{obj}) and (\ref{pmkl}), this is attributed to our new MKL scheme. Note that, KGL outperforms PMKL in several experiments. This is consistent with previous work's claim that MKL may degrade the performance. However, our proposed SMKL can beat KGL in all cases. This also demonstrates the effectiveness of our MKL strategy. With respect to recently developed methods SSR and RMKKM, we also observe considerable improvement. Remember that SSR is based on self-expressiveness in the original space. Compared to traditional methods SC, MKKM, AASC, our advantages become more obvious.  
\subsection{Parameter Analysis}
There are three parameters in our model (\ref{obj}). Figure \ref{yaleacc} shows the clustering accuracy of YALE data set with varying $\alpha$, $\beta$, and $\gamma$. We can observe that the performance is not so sensitive to those parameters. This conclusion is also true for NMI.

\begin{table*}[htbp]
\begin{center}
\tiny
\setlength{\tabcolsep}{1pt}
\renewcommand{\arraystretch}{.9}
\resizebox{.99\textwidth}{!}{
\begin{tabular}{ |l|c|c|c|c|c|c|c|}
\hline
Data &Labeled ($\%$) &GFHF & LGC &S$^3$R&S$^2$LRR& SCAN &SMKL\\
\hline\hline
\multirow{3}{4em}{YALE} & 10 &38.0$\pm$11.91&47.33$\pm$13.96& 38.83$\pm$8.60 &28.77$\pm$9.59& 45.07$\pm$1.30 &\textbf{55.87}$\pm$12.26\\ 
& 30 & 54.13$\pm$9.47&63.08$\pm$2.20& 58.25$\pm$4.25& 42.58$\pm$5.93& 60.92$\pm$4.03&\textbf{74.08}$\pm$1.92\\ 
& 50 & 60.28$\pm$5.16&69.56$\pm$5.42& 69.00$\pm$6.57& 51.22$\pm$6.78 & 68.94$\pm$4.57& \textbf{82.44}$\pm$3.61\\ 
\hline
\multirow{3}{4em}{JAFFE} & 10 & 92.85$\pm$7.76&96.68$\pm$2.76& 97.33$\pm$1.51& 94.38$\pm$6.23& 96.92$\pm$1.68 & \textbf{97.57}$\pm$1.55\\ 
& 30 &98.50$\pm$1.01&98.86$\pm$1.14& 99.25$\pm$0.81& 98.82$\pm$1.05& 98.20$\pm$1.22& \textbf{99.67}$\pm$0.33\\ 
& 50 &98.94$\pm$1.11&99.29$\pm$0.94& 99.82$\pm$0.60& 99.47$\pm$0.59 & 99.25$\pm$5.79& \textbf{99.91}$\pm$0.27\\ 
\hline\hline
\multirow{3}{4em}{BA} & 10 &45.09$\pm$3.09&48.37$\pm$1.98& 25.32$\pm$1.14 &20.10$\pm$2.51&\textbf{55.05}$\pm$1.67& 46.62$\pm$1.98\\ 
& 30 &62.74$\pm$0.92&63.31$\pm$1.03& 44.16$\pm$1.03& 43.84$\pm$1.54&68.84$\pm$1.09&\textbf{68.99}$\pm$0.93\\ 
& 50 &68.30$\pm$1.31&68.45$\pm$1.32& 54.10$\pm$1.55& 52.49$\pm$1.27&72.20$\pm$1.44& \textbf{84.67}$\pm$1.06\\ 
\hline\hline
\multirow{3}{4em}{COIL20} & 10 &87.74$\pm$2.26&85.43$\pm$1.40& \textbf{93.57}$\pm$1.59& 81.10$\pm$1.69&90.09$\pm$1.15 & 91.05$\pm$2.03\\ 
& 30 &95.48$\pm$1.40&87.82$\pm$1.03&96.52$\pm$0.68& 87.69$\pm$1.39 &95.27$\pm$0.93&\textbf{97.89}$\pm$2.00\\ 
& 50 &96.27$\pm$0.71&88.47$\pm$0.45&97.87$\pm$0.10& 90.92$\pm$1.19 &97.53$\pm$0.82& \textbf{99.97}$\pm$0.04\\ 
\hline
\end{tabular}}
\end{center}
\caption{Classification accuracy (\%) on benchmark data sets (mean$\pm$standard deviation). The best results are in bold font.\label{classres}}
\end{table*}

\section{Semi-supervised Classification}
In this section, we assess the effectiveness of SMKL on semi-supervised classification task.
\subsection{Data Sets}
1) \textbf{Evaluation on Face Recognition}: We examine the effectiveness of our graph learning for face recognition on two frequently used face databases: YALE and JEFFE. The YALE face data set contains 15 individuals, and each person has 11 near frontal images taken under different illuminations. Each image is resized to 32$\times$32 pixels. The JAFFE face database consists of 10 individuals, and each subject has 7 different facial expressions (6 basic facial expressions +1 neutral). The images are resized to 26$\times$26 pixels. \\
2) \textbf{Evaluation on Digit/Letter Recognition}: In this experiment, we address the digit/letter recognition problem on the BA database. The data set consists of digits of ``0" through ``9" and letters of capital ``A" to ``Z". Therefore, there are 39 classes and each class has 39 samples.\\
3) \textbf{Evaluation on Visual Object Recognition}: We conduct visual object recognition experiment on the COIL20 database. The database consists of 20 objects and 72 images for each object. For each object, the images were taken 5 degrees apart as the object is rotating on a turntable.  The size of each image is 32$\times$32 pixels.  \\
Similar to clustering experiment, we construct 7 kernels for each data set. They include: four Gaussian kernels with $t$ varies over $\{0.1, 1, 10, 100\}$; a linear kernel $K(x,y)=x^T y$; two polynomial kernels $K(x,y)=(a+x^T y)^2$ with $a\in\{0,1\}$.

\subsection{Comparison Methods}
We compare our method with several other state-of-the-art algorithms.
\begin{itemize}
\itemsep-.2em 
\item {\textbf{Local and Global Consistency (LGC)} \cite{zhou2004learning}: LGC is a popular label propagation method. For this method, kernel matrix is used to compute $L$. }
\item{\textbf{Gaussian Field and Harmonic function (GFHF)} \cite{zhu2003semi}: Different from LGC, GFHF is another mechanics to infer those unknown labels as a process of propagating labels through the pairwise similarity.}
\item{\textbf{Semi-supervised Classification with Adaptive Neighbours (SCAN)} \cite{nie2017multi}: Based on adaptive neighbors method, SCAN shows much better performance than many other techniques.  }
\item{\textbf{A Unified Optimization Framework for SSL} \cite{li2015learning}: Li et al. propose a unified framework based on self-expressiveness approach. By using low-rank and sparse regularizer, they have S$^2$LRR and S$^3$R method, respectively.}
\end{itemize}
\subsection{Performance Evaluation}
We randomly choose 10\%, 30\%, 50\% portions of samples as labeled data and repeat 20 times. Classification accuracy and deviation are shown in Table \ref{classres}. More concretely, for GFHF and LGC, the constructed seven kernels are tested and the best performance is reported. Unlike them, SCAN, S$^2$LRR, S$^3$R, and SMKL, the label prediction and graph learning are conducted in a unified framework.

As expected, the classification accuracy for all methods monotonically increase with the increase of the percentage of labeled samples. As can be observed, our SMKL method outperforms other state-of-the-art methods in most cases. This confirms the effectiveness of our proposed method on SSL task. 

\section{Conclusion}
This paper proposes a novel multiple kernel learning framework for clustering and semi-supervised classification. In this model, a more flexible kernel  learning strategy is developed to enhance the representation ability of the learned optimal kernel and to assign weight for each base kernel. An iterative algorithm is designed to solve the resultant optimization problem, so that graph construction, label learning, kernel learning are boosted by each other. Comprehensive experimental results clearly demonstrates the superiority of our method.
\section*{Acknowledgments}
This paper was in part supported by two Fundamental
Research Funds for the Central Universities of
China (Nos. ZYGX2017KYQD177, ZYGX2016Z003), Grants from the Natural
Science Foundation of China (No. 61572111) and a 985
Project of UESTC (No. A1098531023601041) . 
\bibliographystyle{named}
\bibliography{self}

\end{document}